\documentclass[a4paper]{llncs}
\usepackage{geometry}
\usepackage{amssymb}
\usepackage{graphicx}
\usepackage{multirow}
\usepackage{comment}
\usepackage{amsmath}
\usepackage{url}
\urldef{\mailsa}\path|jay.patravali@qure.ai|
\urldef{\mailsb}\path|shubham.jain@qure.ai|
\urldef{\mailsc}\path|sasank.chilamkurthy@qure.ai|    
\newcommand{\keywords}[1]{\par\addvspace\baselineskip
\noindent\keywordname\enspace\ignorespaces#1}

\begin{document}

\mainmatter  % start of an individual contribution

\title{2D-3D Fully Convolutional Neural Networks for Cardiac MR Segmentation}

\titlerunning{2D-3D Fully Convolutional Neural Networks for Cardiac MR Segmentation}

\author{Jay Patravali, Shubham Jain, 
\and Sasank Chilamkurthy}
\authorrunning{Jay Patravali, Shubham Jain \and Sasank Chilamkurthy }

\institute{Qure.ai\\
\mailsa\\\mailsb\\\mailsc\\
\url{www.qure.ai}}


\toctitle{Lecture Notes in Computer Science}
\tocauthor{Authors' Instructions}
\maketitle

\begin{abstract}
In this paper, we develop a 2D and 3D segmentation pipelines for fully automated cardiac MR image segmentation using Deep Convolutional Neural Networks (CNN). Our models are trained end-to-end from scratch using the ACD Challenge 2017 dataset comprising of 100 studies, each containing Cardiac MR images in End Diastole and End Systole phase. We show that both our segmentation models achieve near state-of-the-art performance scores in terms of distance metrics and have convincing accuracy in terms of clinical parameters. A comparative analysis is provided by introducing a novel dice loss function and its combination with cross entropy loss. By exploring different network structures and comprehensive experiments, we discuss several key insights to obtain optimal model performance, which also is central to the theme of this challenge.

\keywords{Deep Learning, Medical Image Analysis, Computer Vision, MR Segmentation}
\end{abstract}

\section{Introduction}
MR imaging is an effective non-invasive procedure for diagnosis and treatment of known or suspected Cardiac diseases. Cardiac MR images can produce highly detailed pictures of different structures within the heart. Delineation of these structures can provide relevant diagnostic information and evaluate the overall functioning of the heart. Segmentation of left ventricle, right ventricle and the myocardium can be used to calculate relevant diagnostics parameters such as ejection fraction and myocardial mass. Due to massive volumes of cardiac image data, relying on manual delineations can be a time-consuming process, often prone to error and rater variability. Hence there is a critical need for accurate, reproducible and fully-automated methods for cardiac segmentation. 

In recent works, Deep Learning and Convolutional Neural Networks (CNNs) have shown tremendous progress in fully-automated segmentation tasks. The growing success of CNNs in solving computer vision problems such as image recognition and classification \cite{rcnn,resnets} can be  attributed to its ability in learning a hierarchical representation of the input data, without relying on hand-crafted features. Deep learning techniques for segmentation have defined the state-of-the-art using Fully Convolutional Networks (FCN) \cite{fcn}. The idea behind FCN is to use a contracting path to extract features at different spatial scales followed by an expanding path to upsample and increase the spatial resolution of learned features.

For segmentation in medical images, U-Net\cite{unet2d} is a well established 2D CNN architecture that builds upon the FCN. By adding skip connections between the contracting and expanding paths, the U-Net model showed reasonable segmentation accuracy  with very few training samples. Cardiac segmentation based on original FCN have been proposed \cite{tran} with modifications to make it faster and memory efficient \cite{arterys}. As raw 3D MR volumes are fed slice-by-slice as inputs to these 2D CNN models, they fail to capture the spatial contextual information required to segment the whole heart. To that end, U-Net3D\cite{3Dunet} extends the 2D U-Net model by replacing its 2D convolutional operations with its 3D counterparts. In similar way, V-Net\cite{milletari} employs a 3D CNN model with a novel dice loss function showing convincing results in medical image segmentation. To our best knowledge, there are very few methods that have applied 3D CNNs for Cardiac Segmentation and have obtained satisfactory performance.

%\subsection{Automatic Cardiac Delineation Segmentation Challenge 2017}

In our work, we develop a fully-automated 2D and 3D CNN models designed to segment the Left Ventricle, Right Ventricle and Myocardium. This segmentation task is part of the  Automatic Cardiac Detection Challenge 2017 \cite{acdc}. The 2D segmentation model is trained slice-by-slice, whereas we compute volumetric segmentation for the 3D model. Our models are easy to implement, have modular architecture, and relatively short training and testing times. We introduce a new dice loss function, and compare its performance with traditional cross entropy loss and combined cross entropy-dice loss. Through our experiments we also compare and analyze the performance of our 2D and 3D models, both which achieve near state-of-the-art accuracy scores in terms of geometric metrics and  clinical validity.

\section{Method}
\subsection{Network Architecture}

\begin{figure}[tp!]
\centering
\includegraphics[width =0.9\textwidth]{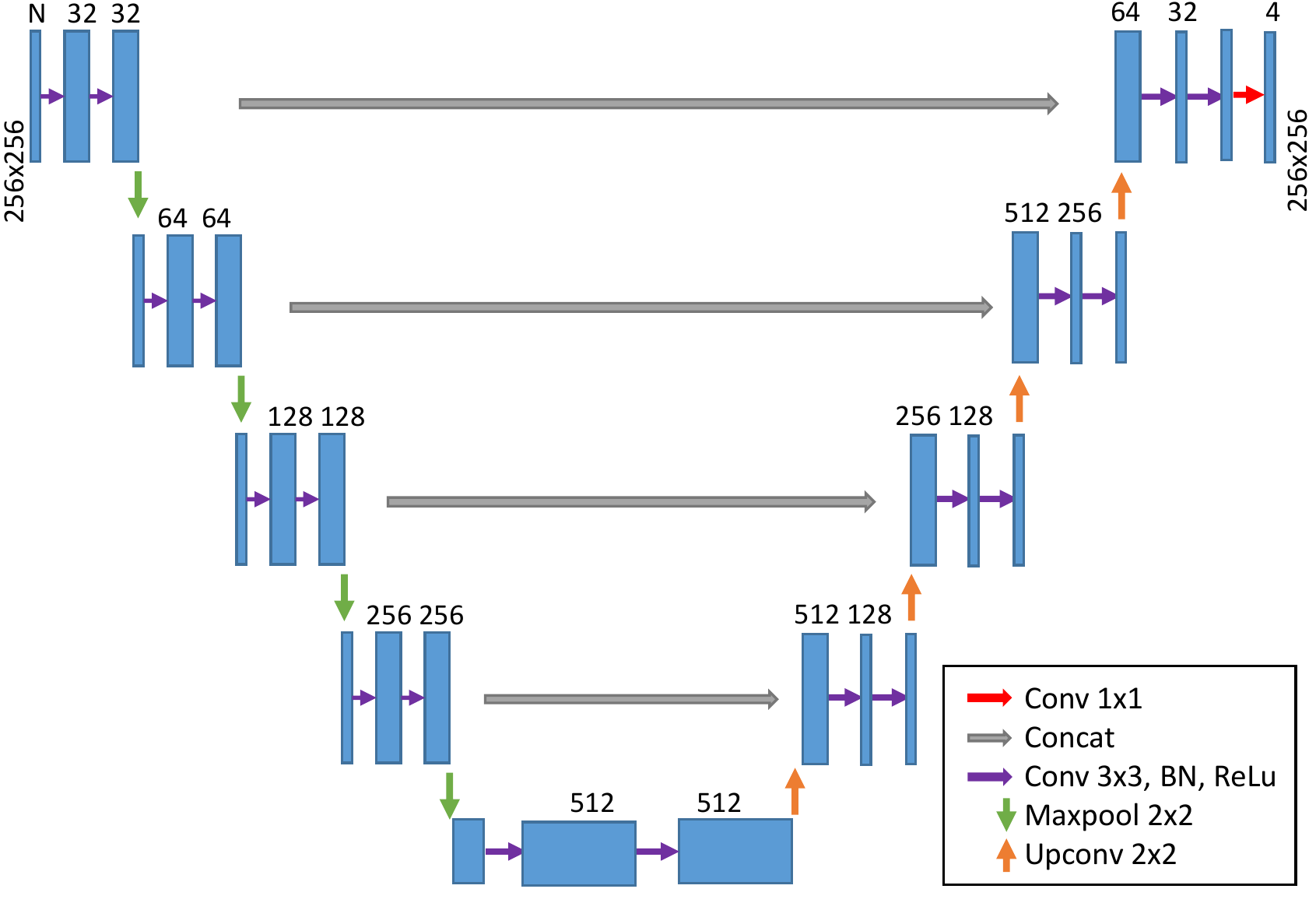}
\caption{2D Model Architecture.}
\label{fig:2d-unet}
\end{figure}

We develop a 2D segmentation model architecture that is adapted from U-Net \cite{unet2d} as illustrated in Fig. \ref{fig:2d-unet}. On left side is the ``contracting" stage and on the right side is``expanding" stage. At the bottom is a base layer. We provide an option to feed varying number (N) of image slices  that can be passed as input channels to the model. Here, N can be 1 for single image slice or more. Every  step on the contracting path consists of a series of a 3x3 convolutions (conv 3x3),  batch normalization (bn) \cite{bn},  rectified linear unit (ReLU)  and conv 3x3 in a sequence that forms a $conv\_bn\_relu$ block. Two of $conv\_bn\_relu$ blocks in succession forms a \textit{Conv} block that doubles the number of feature channels. The contracting path downsamples the image with a 2x2 maxpool operation of stride 2. Similar to contracting stage, every step in expanding stage has a sequence of conv 3x3,  bn,  ReLU  and conv 3x3 in a series of two consecutive blocks.  The images  are upsampled with a 2x2 up-convolution (upconv 2x2) with stride 2. For upsampling of images, a sequence of 2x2 up-convolution (upconv 2x2) with stride 2, concatenations and a \textit{Conv} block forms a \textit{deconv} block. A  final 1x1 convolution layer maps the  32 feature channels to 4 classes. 

\begin{figure}[tp!]
\centering
\includegraphics[width = 0.9\textwidth]{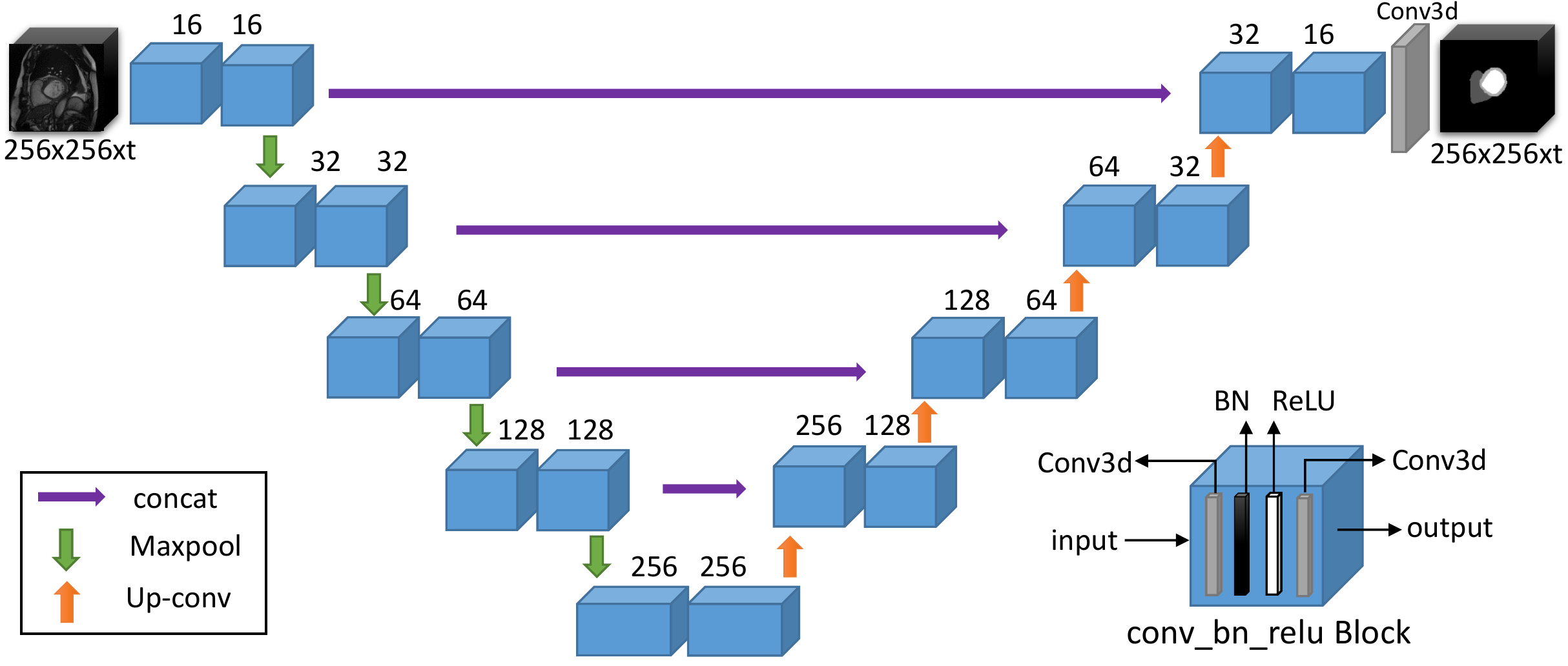}
\caption{3D Model Architecture.}
\label{fig:3d-unet}
\end{figure}

Our 3D model is an extension to our 2D model with few modifications, and finds similarities with U-Net 3d \cite{3Dunet}. First, we replace all 2D operations with its 3D counterparts. Second, due to memory constraints and less number of training examples we limit the maximum number of feature maps to 256. Overall since the number of slices are less across the dataset (9 avg.), we apply 3D 1x2x2 maxpooling operation only in X and Y leaving out the Z dimension. This allows our 3D model to accept input volumes of varying slices at training or inference stage. Similar to our 2D model, every step in contracting and expanding stage consists of two repeating blocks, where each block is a sequence of conv 3x3x3, bn, ReLU, conv 3x3x3. Due to symmetric nature of the model, we can simultaneously add, remove or modify blocks across both paths. Additionally, the size of input data and final outputs images remains the same for both our 2D and 3D models. Thus, we are able to maintain modularity for faster experimentation and modification at the time of training and design. To prevent overfitting, we add dropout layers \cite{dropout} with probability values of 0.5 in the last and 0.3 in the second last layer of the contracting stage, in both 2D and 3D models.

\subsection{Dataset, Preprocessing and Augmentation}
Our models are trained end-to-end from scratch using MICCAI's ACD Challenge 2017 dataset. It contains 150 exams of fully-annotated cardiac MRI's. Out of these 100 are used for training phase and 50 for testing phase of this challenge. These exams are obtained from multiple patients, each consisting of scans from End Diastole and End Systole phase taken in short axis orientation. 

Since the data acquisition can bring inconsistencies in dataset, its necessary to carry preprocessing steps to ensure that the model receives uniform inputs. To remove noise and enhance contrast, we use contrast limited adaptive histogram localization (CLAHE) \cite{clahe}. Then, we normalize the intensity values of all images between the range of 1-99 percentiles. Finally, we clip the image pixel values between 0 and 1. 

To ensure both 2D and 3D model perceive heart features in similar proportion, we do a resampling operation across all  input volumes to a common voxel spacing of 1.5x1.5x10 mm. For 2D segmentation we resize and crop the images to fixed size of 256x256. Due to maxpool operation applied only on height and width in our 3D segmentation model, when testing the 3D model, we can feed it image volumes of varying number of slices. At training, the 3D model is fed with raw input volumes that are resized and cropped to 256x256x12.

We apply a light data augmentation techniques on-the-fly to efficiently feed the input data volumes into our model. On a random basis, the data is rotated between -15 to +15 degree, and scaled between 0.9 - 1.1 range. This ensures slight robustness and variability in training the network.

\subsection{Training}
We train the 2D segmentation model by feeding it raw input images slice-by-slice. Whereas the 3D segmentation is trained by feeding it with entire 3D input volumes. Both the 2D and 3D models are trained with different optimization functions described more in detail in the following subsection. 

During the training process, weights are updated using  stochastic gradient descent with a momentum of 0.99. The initial learning rate is decayed by a factor of 10 every 30 epochs. For the training phase of this challenge, we do a 5-fold cross validation, which leaves 80 patients for training and 20 for validation. Training is completed after 300 epochs. Best models are checkpointed and stored for testing. For 2D segmentation we use a batch size of 8, whereas for 3D segmentation we use a batch size of 4.  

Our models are implemented entirely using the PyTorch \cite{pytorch} framework, due to its flexibility in experimentation. We run all experiments on a standard workstation equipped with 64 GB of memory, Intel(R) Core(TM)  i7-6700K CPU clocking at 4.00GHz, with a 12 GB NVidia Titan X Pascal GPU. 

%%%%%%%%%%%%%%%%%%%% WIP out %%%%%%%%%%%%%%%%%%%%%

\subsection{Optimization Function}
In this section we describe three optimization functions that are used for training our 2D and 3D segmentation models. We use these functions to compare the performance of 2D and 3D models. At training, we apply a pixel-wise softmax activation in the final layer of the model to get the predicted probabilities $p(x, i)$ for each class $i$ at each pixel $x$. The targets at location $x$ is denoted by $t(x)$.

\subsubsection{Cross Entropy Loss}
In segmentation tasks, the standard practice is to apply cross entropy loss function to measure the pixel-wise probability error between the predicted output and target and sum the errors across all the pixels.
In addition to this, we apply weights to each class ($w_i$ for class $i$) to offset the imbalance of pixel frequency across different classes. Concretely, the weighted cross entropy loss $L_{\text{CE}}$ is defined as,

\begin{equation}
L_{\text{CE}} = - \sum_{x} w_{t(x)} \log(p(x, t(x)))
\label{eq:pixelce}
\end{equation}

\subsubsection{Dice Loss}
The Dice's Coefficient is a metric to measure the similarity between two given samples. Extending it as a loss function as shown in \cite{milletari}, improves the performance when dealing with situations where background pixels are higher than the labels. Here, we introduce a novel dice loss  
$L_{\text{dice}}$ that is a logarithmic value of the dice score, making it easier to optimize.
Similar to weighted cross entropy, we use weights to offset the class imbalance.
Dice loss $L_{\text{dice}}$ is weighted sum of dice losses $l_i$ for each class $i$ is given as,

\begin{equation}
L_{\text{dice}} = \sum_i w_i l_i
\end{equation}

For class $i$, let's denote its binary map by $t_i$. i.e,

\begin{equation}
t_i(x)=
\begin{cases}
  1, & \text{if}\ t(x) = i \\
  0, & \text{otherwise}
\end{cases}
\end{equation}

Then, dice loss for class $i$ is given by,

\begin{equation}
l_i = \log\left(2 - \frac{\sum_x t_i(x)p(x, i) + \epsilon}{\sum_x t_i(x) + p(x, i) + \epsilon}\right)
\end{equation}

% \begin{equation}
% DiceLoss(y, class) = \sum_{class} weights(class) * dicelog(y, class)
% \label{eq:quantization_eqn}
% \end{equation}
% \begin{equation}
% dicelog(y, class) = log(2 - \frac{y * y(class)}{|y(class)| + |y|})
% \label{eq:quantization_eqn}
% \end{equation}

\subsubsection{Combined Dice Cross Entropy Loss}
While cross-entropy loss optimizes for pixel-level accuracy, the Dice loss function enhances the segmentation quality. Combining these two objective functions, we define a weighted average of cross entropy $L_{\text{CE}}$ and dice loss function $L_{\text{dice}}$ formulated as Cross Entropy-Dice Loss $L_{\text{CE + dice}}$ in Eqn 5. Here,  $\lambda_{\text{CE}}$ and $ \lambda_{\text{dice}}$ are weight parameters for cross entropy loss and Dice loss function respectively.

\begin{equation}
L_{\text{CE + dice}} = 
\lambda_{\text{CE}} * L_{\text{CE}} 
+ 
\lambda_{\text{dice}} * L_{\text{dice}}
\end{equation}

%%%%%%%%%%%%%%%%%%%% Commented out %%%%%%%%%%%%%%%%%%%%%

\section{Results}
\begin{figure}[tp!]
%\centering
\includegraphics[width=\textwidth, keepaspectratio]{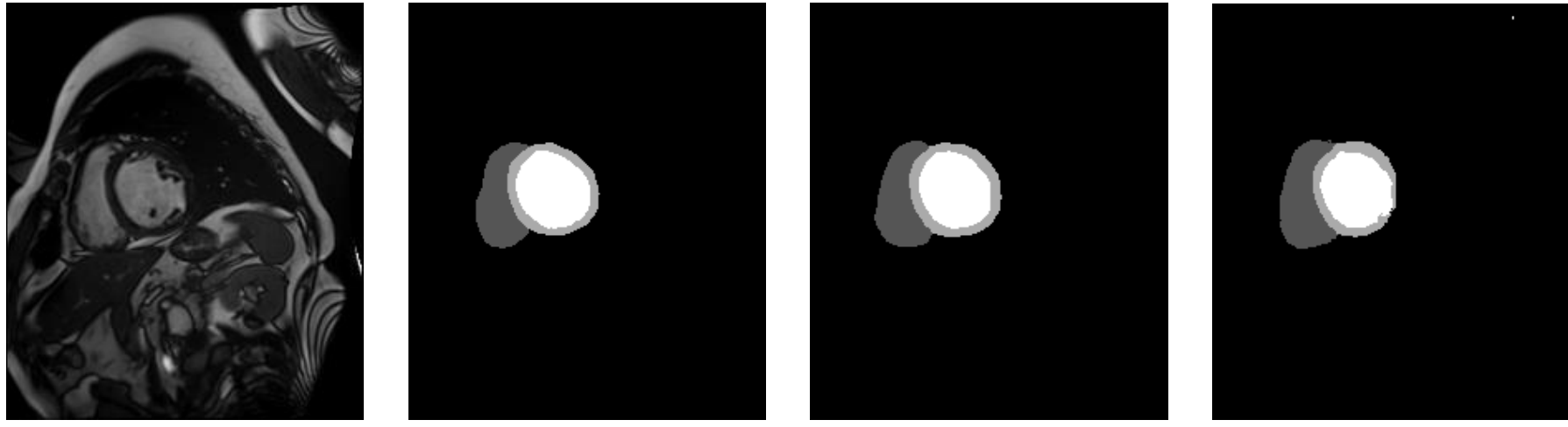}
\caption{Segmentation Results for 2D and 3D model. \textbf{From Left to Right:} Raw MR input image slice, Corresponding ground truth annotation, output predictions from 2D segmentation model and output predictions from 3D segmentation model.}
\label{fig:results1}
\end{figure}

\begin{figure}[tp!]
%\centering
\includegraphics[width=\textwidth,height=\textheight,keepaspectratio]{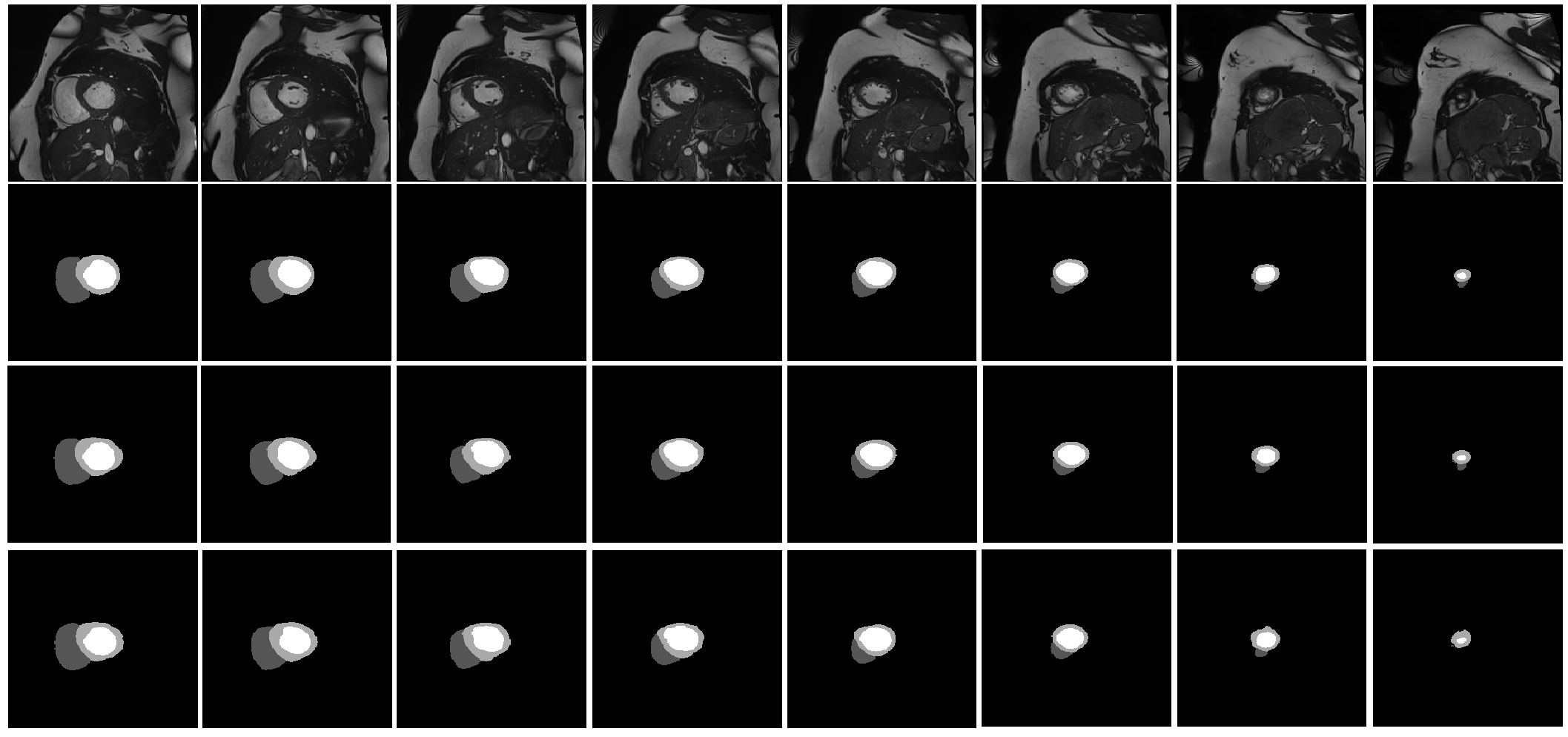}
\caption{Segmentation Results for a full MR image (Complete Phase) from slices 0-8 . \textbf{First Row:} Raw MR input images  \textbf{Second Row:} Corresponding ground truth annotations. \textbf{Third Row:} Output predictions from 2D segmentation model. \textbf{Fourth Row:} Output predictions from 3D segmentation model.}
\label{fig:results2}
\end{figure}

In this section we evaluate the performance of the proposed 2D and 3D segmentation models in terms of geometric or distance metrics  and clinical metric scores for all 100 studies  provided in the training phase of ACDC 2017 contest. Table \ref{2d-seg-dist} and Table \ref{3d-seg-dist} presents the distance metric scores for our 2D model and 3D model respectively. For distance metric, we utilize the Dice Score and the  Hausdorff Distance to measure the accuracy of segmented Left Ventricle (LV), Right Ventricle (RV) and Myocardium (MYO) in both end diastole (ED) and end systole (ES) phase. For each model, we compare the performance of the three optimization functions namely, the Cross Entropy Loss (CE Loss), Dice Loss and Combined Cross Entropy-Dice Loss (Dice-CE Loss). We observe that our proposed dice loss function outperforms CE Loss and CE-Dice Loss functions across all metrics in both 3D and 2D models. The Hausdorff distances in 3D models is observed to be much higher than 2D models, due to false positives as far-off speckles in 3D space. Illustration of results for both 2D and 3D models are provided in Fig. \ref{fig:results1} and Fig. \ref{fig:results2}. Overall, we achieve near equal distance metric scores when compared to \cite{clement}.

For clinical metrics, we use Correlation Coefficient (CC), Bias and Limits of Agreement (LOA). Our clinical metric results for 2D and 3D models are presented in Table \ref{2d-seg-clin} and \ref{3d-seg-clin} respectively. For both 2D and 3D models, the performance using cross-entropy and dice-loss functions for LV and RV is fairly similar, however the difference in performance is significant for MYO where dice-loss outperforms cross-entropy optimization. While distance metric scores are fairly similar for both 2D and 3D model, in clinical parameters we observe that the 3D model outperforms the 2D model. 

Although accuracy scores are important when making clinical decisions, run-time efficiency and memory usage of the algorithm are also crucial to apply it in real-world applications. Our 2D model takes 2.9 hours to train using 4GB of GPU memory. At testing, it takes 0.3s and 1.2GB GPU memory to generate a single output(whole phase). Whereas our 3D model requires 2.6 hours for training and 4GB of GPU memory. At test time, it can generate output within 0.3s using 2GB GPU memory. This shows that our 2D and 3D models are efficient to train, are light-weight and relatively easy to deploy in clinical settings.

\begin{table}[tp!]
\centering
\caption{2D Segmentation: Distance Metric Results}
\label{2d-seg-dist}
\begin{tabular}{|l|c|c|c|c|c|c|c|c|c|c|c|c|}
\hline
\multirow{3}{*}{} & \multicolumn{6}{c|}{Dice Score}                                                         & \multicolumn{6}{c|}{Hausdorff Distance}                                                          \\ \cline{2-13} 
                  & \multicolumn{2}{c|}{LV}       & \multicolumn{2}{c|}{RV} & \multicolumn{2}{c|}{MYO}      & \multicolumn{2}{c|}{LV}       & \multicolumn{2}{c|}{RV}        & \multicolumn{2}{c|}{MYO}        \\ \cline{2-13} 
                  & ED            & ES            & ED              & ES    & ED            & ES            & ED            & ES            & ED            & ES             & ED             & ES             \\ \hline
CE Loss           & 0.95          & 0.90          & 0.87            & 0.76  & 0.79          & 0.82          & 13.92         & 17.67         & 27.40         & 27.73          & 23.81          & 22.11          \\ \hline
Dice Loss         & \textbf{0.95} & \textbf{0.90} & \textbf{0.90}   & 0.79  & \textbf{0.86} & \textbf{0.88} & 9.51          & 12.29         & 16.1          & 20.38          & \textbf{13.45} & \textbf{14.88} \\ \hline
Dice-CE Loss & 0.95          & 0.90          & 0.89            & \textbf{0.81}  & 0.83          & 0.84          & \textbf{9.15} & \textbf{11.7} & \textbf{16.0} & \textbf{18.22} & 13.87          & 15.35          \\ \hline
\end{tabular}
\end{table}

\begin{table}[tp!]
\centering
\caption{3D Segmentation: Distance Metric Results}
\label{3d-seg-dist}
\begin{tabular}{|l|c|c|c|c|c|c|c|c|c|c|c|c|}
\hline
\multirow{3}{*}{} & \multicolumn{6}{c|}{Dice Score}                                                         & \multicolumn{6}{c|}{Hausdorff Distance}                                                             \\ \cline{2-13} 
                  & \multicolumn{2}{c|}{LV}       & \multicolumn{2}{c|}{RV} & \multicolumn{2}{c|}{MYO}      & \multicolumn{2}{c|}{LV}         & \multicolumn{2}{c|}{RV}         & \multicolumn{2}{c|}{MYO}        \\ \cline{2-13} 
                  & ED            & ES            & ED              & ES    & ED            & ES            & ED             & ES             & ED             & ES             & ED             & ES             \\ \hline
CE Loss           & 0.94          & 0.89          & 0.86            & 0.73  & 0.76          & 0.81          & 12.36          & 14.41          & 25.85          & 29.57          & 43.47          & 43.82          \\ \hline
Dice Loss         & \textbf{0.95} & \textbf{0.90} & \textbf{0.91}   & \textbf{0.83}  & \textbf{0.85} & \textbf{0.86} & 14.95          & 14.35          & \textbf{23.15} & \textbf{22.14} & \textbf{37.75} & \textbf{38.50} \\ \hline
Dice-CE Loss & 0.94          & 0.89          & 0.91            & 0.81  & 0.83          & 0.85          & \textbf{10.71} & \textbf{11.52} & 38.01          & 32.26          & 43.28          & 44.98          \\ \hline
\end{tabular}
\end{table}

\begin{table}[tp!]
\centering
\caption{2D Segmentation: Clinical Metric Results}
\label{2d-seg-clin}
\begin{tabular}{|l|c|c|c|c|c|c|c|c|c|}
\hline
\multirow{3}{*}{} & \multicolumn{6}{c|}{Ejection Fraction}                                                                           & \multicolumn{3}{c|}{Myocardial Mass}                    \\ \cline{2-10} 
                  & \multicolumn{3}{c|}{LV}                                 & \multicolumn{3}{c|}{RV}                                & \multicolumn{3}{c|}{MYO}                                \\ \cline{2-10} 
                  & CC             & Bias           & LOA                   & CC             & Bias          & LOA                   & CC             & Bias           & LOA                   \\ \hline
CE Loss           & \textbf{0.95} & -0.74          & \textbf{-12.48,11.00} & 0.822          & 9.79          & \textbf{-10.89,30.47} & 0.93          & -43.85         & -92.12,4.42           \\ \hline
Dice Loss         & 0.88          & 1.06           & -18.10,20.22          & \textbf{0.822} & \textbf{9.35} & -11.67,30.37          & \textbf{0.95} & \textbf{-6.32} & \textbf{-39.46,26.82} \\ \hline
Dice-CE Loss & 0.93          & \textbf{-0.46} & -15.67,14.75          & 0.813          & 5.66          & -16.59,27.91          & 0.94          & -29.31         & -68.20,9.58           \\ \hline
\end{tabular}
\end{table}

\begin{table}[tp!]
\centering
\caption{3D Segmentation: Clinical Metric Results}
\label{3d-seg-clin}
\begin{tabular}{|l|c|c|c|c|c|c|c|c|c|}
\hline
\multirow{3}{*}{} & \multicolumn{6}{c|}{Ejection Fraction}                                                                          & \multicolumn{3}{c|}{Myocardial Mass}                    \\ \cline{2-10} 
                  & \multicolumn{3}{c|}{LV}                               & \multicolumn{3}{c|}{RV}                                 & \multicolumn{3}{c|}{MYO}                                \\ \cline{2-10} 
                  & CC             & Bias          & LOA                  & CC             & Bias          & LOA                    & CC             & Bias          & LOA                    \\ \hline
CE Loss           & \textbf{0.975} & \textbf{1.04} & \textbf{-7.67,9.75} & 0.756          & 9.62          & -15.11,34.35          & 0.922          & -48.17        & -97.51,1.17           \\ \hline
Dice Loss         & 0.956          & 1.51          & -10.09, 13.11        & 0.825          & \textbf{4.99} & -17.17,27.15          & \textbf{0.958} & \textbf{3.77} & \textbf{-24.91,32.45} \\ \hline
Dice-CE Loss & 0.956          & 1.25          & -10.37,12.87        & \textbf{0.867} & 6.19          & \textbf{-12.09,24.47} & 0.950          & -10.08        & -42.85,22.69          \\ \hline
\end{tabular}
\end{table}

\section{Discussion}

%We addressed this by exploring 3d models for cardiac segmentation and dice-loss for optimization.

\subsubsection{Model Structures}
Since 2D segmentation model  is trained by splitting MR volumes into slices, it lacks the spatial context across the 3D volume. This reduces the model's performance for slices at end of the phase, where the ratio of heart structure to background pixels is less. To solve this, we design our 2D model to accept a stack of image slices as input channels where the output is predicted for the middle slice. Given this design, we have 3 input options to pass as inputs: 1, 3 and 5. Among the three, we obtain the best performance with 3-input slices and report the scores for it in our results section.

Analyzing the 3D CNN segmentation model, we observe that the 3D model doesn't meet the expectations in performance improvement over 2D, given its ability to exploit 3D structure from input volumes. We see that Dice Score for RV in 3D is better 2D model given the fact that it has complex shape and intensity inhomogeneities. Thus, predicting RV using single slice is much more difficult compared to looking at complete 3D context. Further improvements in performance can be brought about using post-processing techniques. Due to the modularity of 3D model architecture, we were able to quickly explore several designs and concepts. For example, we tried the recently introduced subpixel CNN's \cite{sr} that proposes using subpixel layers as opposed to transposed convolutions. We executed this by replacing the \textit{deconv} blocks in the expanding paths with a $subpixel$ block that comprises of a $subpixel$ layer between two  $conv\_bn\_relu$ blocks. However, no performance improvements were to be observed.

\subsubsection{Data Augmentation}
At the initial stages of model design and training, we applied variety of data augmentation techniques demonstrated in \cite{milletari,3Dunet},  to incorporate randomness and robustness in the training the network. These include elastic deformations, random intensity jitter and affine transformations that include rotation, shearing, translation and flipping. Acquiring sub-par performances at testing, we found that applying heavy augmentations might misrepresent the anatomical structure of the heart. Instead, we opt to apply light data augmentations consisting of random rotations and scaling that can naturally match the variability of taking MR scans in real-world settings. With this modification, we observed a jump in performance scores for both 2D and 3D segmentation model.
\vspace{-0.1cm}
\subsubsection{Optimization Functions}
Using our Dice loss as objective function improves the performance significantly for 2D and 3D models. As compared to pixel-level error optimization, dice loss is more robust and better at capturing the spatial context over the entire image. As shown in Table 1 and 2, the best dice scores are achieved by using dice loss function.
\vspace{-0.1cm}
%\subsubsection{Performance}

%% Here we talk about the problems in cleaning data, different methods we tried, why X works and not y

\section{Conclusion}
This paper introduces a  2D and 3D convolutional neural network for fully-automated cardiac MR segmentation. Our models have light-weight modular architecture, easy implementation and run-time efficiency.  Using multiple loss criterion, we compare and analyze the performance of 2D and 3D model pipelines and show that both our models achieve near  state-of-the-art accuracy scores in terms of distance metrics. With convincing performance in clinical accuracy metrics, we also prove our model's viability in real-world practical applications. Through our discussions, we derive several insights that can be used for optimizing overall performance of these segmentation models. For future work, we plan to utilize our segmentation models to learn and classify different cardiac diseases.

\end{document}